\documentclass[10pt,twocolumn,letterpaper]{article}

\usepackage{iccv}
\usepackage{times}
\usepackage{epsfig}
\usepackage{graphicx}
\usepackage{amsmath}
\usepackage{amssymb}

\usepackage{booktabs}
\usepackage{multirow}
\usepackage{bbm}
\usepackage{pifont}
\usepackage[ruled,vlined, noend]{algorithm2e}
\usepackage{xcolor}
\definecolor{citecolor}{HTML}{0071bc}

\usepackage[pagebackref,breaklinks,colorlinks,anchorcolor=blue,citecolor=citecolor,pagebackref=true,breaklinks=true,letterpaper=true,colorlinks,bookmarks=false]{hyperref}

\usepackage[capitalize]{cleveref}
\crefname{section}{Sec.}{Secs.}
\Crefname{section}{Section}{Sections}
\Crefname{table}{Table}{Tables}
\crefname{table}{Tab.}{Tabs.}

\newlength\savewidth

\usepackage{adjustbox}
\usepackage{tabularx}
\usepackage{xcolor}
\usepackage{colortbl}

\usepackage[pagebackref=true,breaklinks=true,letterpaper=true,colorlinks,bookmarks=false]{hyperref}

\iccvfinalcopy 

\def\iccvPaperID{5555} 

\ificcvfinal\fi

\begin{document}

\title{Augment and Criticize: Exploring Informative Samples for Semi-Supervised \\ Monocular 3D Object Detection}

\author{
Zhenyu Li$^1$\thanks{This work was done when Zhenyu was an intern at DiDi.}, Zhipeng Zhang$^2$, Heng Fan$^3$, Yuan He$^2$, Ke Wang$^2$, Xianming Liu$^1$, Junjun Jiang$^1$\thanks{Corresponding author (jiangjunjun@hit.edu.cn).}\\
$^1$Harbin Institute of Technology, $^2$DiDi Chuxing, $^3$University of North Texas\\
}

\maketitle
\ificcvfinal\thispagestyle{empty}\fi


\begin{abstract}

    In this paper, we improve the challenging monocular 3D object detection problem with a general semi-supervised framework. Specifically, having observed that the bottleneck of this task lies in lacking reliable and informative samples to train the detector, we introduce a novel, simple, yet effective `Augment and Criticize' framework that explores abundant informative samples from unlabeled data for learning more robust detection models. In the `Augment' stage, we present the Augmentation-based Prediction aGgregation (APG), which aggregates detections from various automatically learned augmented views to improve the robustness of pseudo label generation. Since not all pseudo labels from APG are beneficially informative, the subsequent `Criticize' phase is presented. In particular, we introduce the Critical Retraining Strategy (CRS) that, unlike simply filtering pseudo labels using a fixed threshold (\textit{e.g.,} classification score) as in 2D semi-supervised tasks, leverages a learnable network to evaluate the contribution of unlabeled images at different training timestamps. This way, the noisy samples prohibitive to model evolution could be effectively suppressed. To validate our framework,  we apply it to MonoDLE~\cite{ma2021monodle} and MonoFlex~\cite{zhang2021monoflex}. The two new detectors, dubbed 3DSeMo$_{\text{DLE}}$ and 3DSeMo$_{\text{FLEX}}$, achieve state-of-the-art results with remarkable improvements for over 3.5\% $AP_{3D/BEV} (Easy)$ on KITTI, showing its effectiveness and generality. Code and models will be released.

\end{abstract}
\section{Introduction}

\begin{figure}[!t]
    \centering
    \footnotesize
    \includegraphics[width=\linewidth]{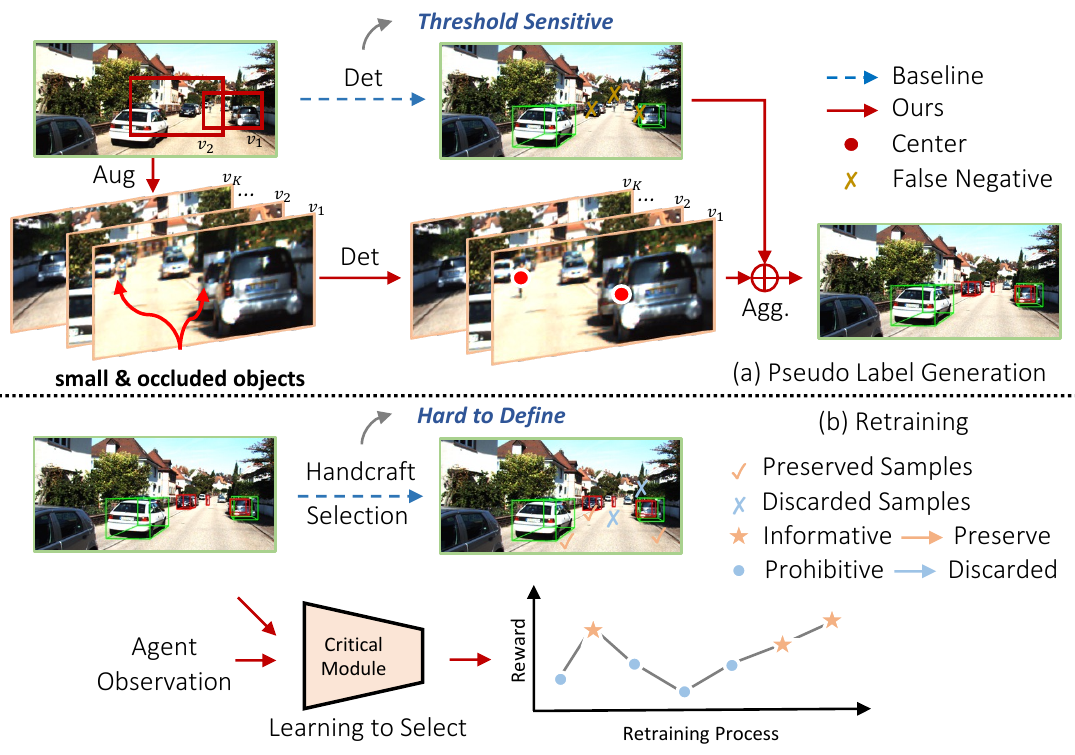}
    \caption{\textbf{Motivation and Proposal.} The differences between our method (\textcolor{red}{red}) and previous semi-supervised learning framework (\textcolor[RGB]{61,145,64}{green}) in pseudo label (PL) generation and student model retraining. The introduced framework can improve detection recall by observing different views of an image (\textcolor{red}{red} dots in (a)), and dynamically determine when to discard an unlabeled sample during training (line-chart in (b)) by the learnable critical module.} 
    \label{fig::teaser}
\end{figure}

Monocular 3D (Mono3D) object detection is an essential and economical way for intelligent agents to perceive 3D information in the real world. However, since the depth information is inevitably lost during the 3D-to-2D projection process~\cite{eigen2014depth}, identifying 3D objects under pure monocular premise is ill-posed in the first place. Fortunately, the recent triumphs in deep learning shed light on circumventing the complex mathematical optimization~\cite{math} and tackling this challenging problem in a data-driven manner~\cite{liu2020smoke,ma2021monodle,zhang2022monodetr,zhang2021monoflex,wang2021fcos3d}. Despite of the improvements, state-of-the-art
Mono3D models still lag far behind human perception capabilities. On the contrary, deep learning has surpassed human-level performance on image classification tasks~\cite{he2015delving}.
One possible reason behind such a performance gap is the notable dataset volume differences: for example, Mono3D models are trained with 3.7K images from KITTI~\cite{geiger2012kitti}, while classification models are trained with 1.2 million images from ImageNet~\cite{deng2009imagenet}.



Why not just simply scale up the dataset volume then? It turns out that 3D annotations are more expensive than 2D ones. High-quality Mono3D datasets need LiDARs for 3D ground truth. Data collection needs carefully calibrated and synchronized different sensors. Taking this into consideration, we believe a data-driven but annotation-efficient approach is necessary to bring the performances of modern Mono3D models to the next level. 

Naturally, semi-supervised learning, which can absorb knowledge from \textit{limited} annotated samples meanwhile exploit beneficial information from the \textit{enormous} unlabeled data, becomes a reasonable choice for this problem. Surprisingly, it has been rarely explored in monocular 3D detection, despite its success in numerous 2D vision tasks~\cite{tarvainen2017meanteacher,berthelot2019mixmatch,sohn2020fixmatch,xie2020self,xu2021softteacher,zhou2022denseteacher,zhang2022semi,yang2022st++,liu2022perturbed,yuan2022semi}. We suspect that the noise in the low-quality pseudo labels during training eventually overwhelms the benefits brought by abundant extra data. Thus, how to effectively find reliable and informative samples from unlabeled data, is crucial to apply semi-supervised learning in Mono3D tasks. In particular, 
two challenges are faced: \emph{How to robustly generate high-quality pseudo labels from unlabeled data} \textbf{and} \emph{How to properly leverage these pseudo labels for effective learning}.

For the first challenge,
prior works in 2D tasks~\cite{xu2021softteacher,chen2022labelmatch} alleviate the pseudo label quality issue by handpicking a threshold to filter putative detection boxes from unlabeled data (see \cref{fig::teaser}). Such strategy might not work well in the Mono3D setting. Firstly, generating precise pseudo labels from images alone is already very difficult (notice the gap between vision-only and multi-modal 3D detection results on public benchmarks); secondly, using a handcrafted selection strategy may overfit to a specific model, reducing the resilience of semi-supervised models. Thus, 
\textbf{\emph{a robust pseudo label generation strategy is needed to tackle the first challenge.}} To this end, we introduce a simple yet effective Augmentation-based Prediction aGgregation strategy, dubbed \textbf{APG}, that aims at \textit{robustly} generating pseudo labels for unlabeled data. The core idea is to aggregate predictions from different observations of an image, which we find effectively reduces the detection biases and improves the robustness of pseudo label generation (see \cref{fig::teaser}). 

For the second challenge, since not all the pseudo labels are beneficially informative (even for the proposed APG), classification scores are usually used in 2D tasks~\cite{chen2022labelmatch} to adapt the pseudo detection boxes for training. However, it ignores that the contribution of each sample during model training should vary across training iterations and the model’s parameters. Therefore, \textbf{\emph{a more adaptive mechanism is needed to guide the training on unlabeled data.}} For that, we propose a `Criticize' module for the second challenge. More specifically, in this stage, a Critical Retraining Strategy (\textbf{CRS}) is imposed to \textit{adaptively} update the model with noisy pseudo labels. In particular, CRS contains a memory bank to preserve evaluation images and a criticize module to determine which pseudo label benefits more when updating the model.
Given a pseudo label, the optimization loss of the model roughly indicates the benefits this particular pseudo label would bring, should this pseudo label be used for back-propagation.
The `Criticize' module discards the pseudo label if the model would be updated towards unsatisfactory directions. As training proceeds, the Criticize module can better adapt to the model optimization progress through cyclical updating of the memory bank.



To summarize, we propose a novel `Augment and Criticize' framework to approach the two challenges in semi-supervised Mono3D object detection. Results on different methods and benchmarks show remarkable improvement. Our contributions are summarized as follow,

\noindent
(\textbf{1}) \emph{We propose a novel `Augment and Criticize' framework for semi-supervised Mono3D object detection.}

\noindent
(\textbf{2}) \emph{We propose an augmented-based prediction aggregation to improve the quality of pseudo labels for unlabeled data.}

\noindent
(\textbf{3}) \emph{We propose a critical retraining strategy that adaptively evaluates each pseudo label for effective model training. }

\noindent
(\textbf{4}) \emph{We integrate our semi-supervised framework into different methods, and results evidence its effectiveness. }

\section{Related Work}


\begin{figure*}[!t]
    \centering
    \footnotesize
    \includegraphics[width=0.98\linewidth]{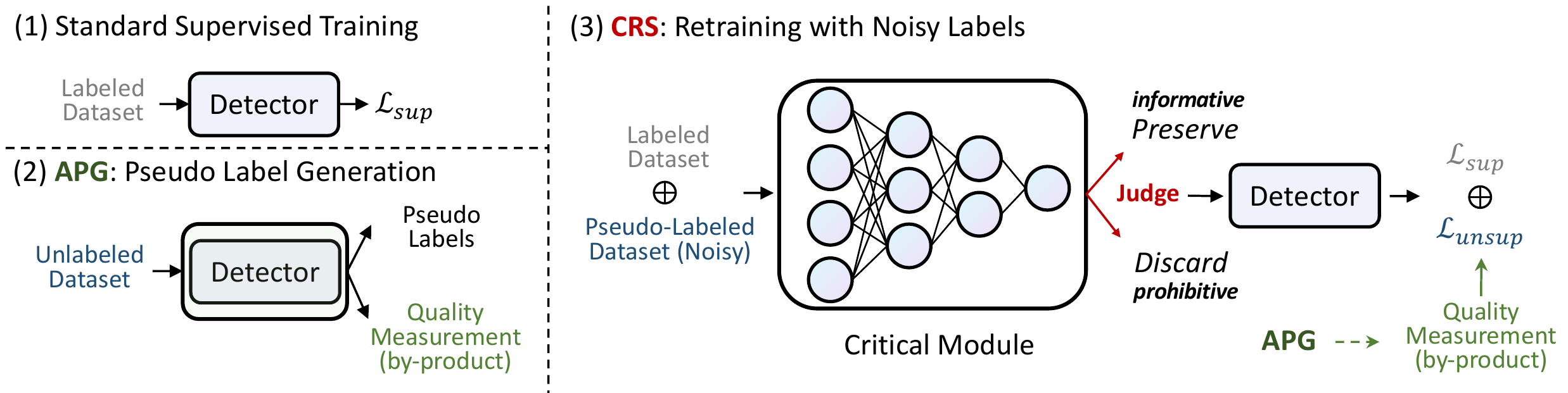}
    \caption{\textbf{`Augment and Criticize' Framework.} We present the three major steps in our semi-supervised scheme, including standard supervised training, pseudo label generation with APG, and retraining with CRS.} 
    \label{fig::overall}
\end{figure*}


\subsection{Monocular 3D Object Detection}
3D object detection is a fundamental task for agents to perceive the surrounding 3D world~\cite{arnold2019survey}. Benefiting from the ubiquitous availabilities of cameras, Mono3D methods have great potential for wide real-world deployment, thus have received extensive attention recently~\cite{brazil2020kinematic,ding2020d4lcn,zhang2021monoflex,shi2021monorcnn,ma2021monodle,lu2021gup,huang2022monodtr}.
Earlier attempts devoted massive efforts to the ill-posed depth estimation problem. By adopting an isolated depth model~\cite{weng2019plpc} to generate pseudo point cloud or lifting 2D features to 3D space~\cite{roddick2018orthographic}, 3D detectors can be applied on such pseudo point cloud or 3D features for object identification. Despite promising results, the hefty computation overhead entailed by dense depth estimation prohibits such methods from practical applications.
By moving depth estimation into an auxiliary head, \cite{liu2020smoke,ma2021monodle,zhang2021monoflex,wang2021fcos3d} enabled end-to-end model training with a neater framework. These methods simultaneously predict object centers in the 2D images and their corresponding depth in the 3D space. Object 3D locations and dimensions can then be easily recovered with camera calibration.
Representative methods like SMOKE~\cite{liu2020smoke} and MonoDLE~\cite{ma2021monodle} adopt CenterNet-like architectures~\cite{zhou2019centernet}, whereas FCOS3D~\cite{wang2021fcos3d} and PGD~\cite{wang2022pgd} extend the 2D FCOS detector~\cite{tian2019fcos} into a 3D detection model.
In this paper, we aim to design a general semi-supervised framework, which is agnostic to specific model designs, to push the envelope of modern Mono3D object detectors.

\subsection{Semi-Supervised Learning}

Semi-Supervised Learning (SSL) is attractive because of its capability to further unveil the power of machine learning with abundant cheap unlabeled data~\cite{van2020survey,van2020survey,radford2015unsupervised,lee2013pseudo,berthelot2019mixmatch,sohn2020fixmatch,rasmus2015semi}. Due to the space limitation, this section only reviews self-training-based methods, which is one of the most engaging directions in SSL~\cite{mclachlan1975iterative, scudder1965probability}. In general, self-training-based semi-supervised learning methods first train a teacher model with a small set of human-annotated data. The teacher model then generates pseudo labels on a much larger set of unlabeled data.
Finally, a student model is trained with both human-labeled and self-annotated data.
Such a paradigm has demonstrated great success in 
image classification~\cite{tarvainen2017meanteacher,berthelot2019mixmatch,sohn2020fixmatch,xie2020self}, semantic segmentation~\cite{yang2022st++,liu2022perturbed,yuan2022semi}, and 2D object detection~\cite{xu2021softteacher,zhou2022denseteacher,zhang2022semi}.
While different applications usually require additional bells and whistles, the core components of semi-supervised learning remain unchanged: how to generate high-quality pseudo-label, and how to retrain student models effectively.
Mean-Teacher~\cite{tarvainen2017meanteacher} proposes temporal ensembling to facilitate retraining. Soft-Teacher~\cite{xu2021softteacher} utilizes the classification score to reweight loss and imposed 2D box jitter to filter unreliable pseudo labels. ST++~\cite{yang2022st++} adopts strong augmentations on the unlabeled samples and leverages evolving stability during training to prioritize high-quality labels.
Compared with well-studied 2D tasks, it is much more challenging for Mono3D detection to collect reliable pseudo labels. Although such issue can be alleviated by introducing multi-view consistency~\cite{lian2022semimono3d}, compared with abundant single-view datasets, high-quality stereo or multi-view datasets are much harder to collect (device-wise multi-view). Besides, learning consistency among video frames is vulnerable to moving objects (temporal-wise multi-view). Thus we believe, it is the single-view scenario that semi-supervised learning can make the most impact. In this paper, we focus on the design of effective semi-supervised learning frameworks for Mono3D object detection.



%



\section{Method}
Our proposed semi-supervised framework (\cref{fig::overall}) is detailed in this section.
We first recap the definition of semi-supervised Mono3D object detection task and introduce the vanilla self-training scheme in~\cref{Preliminary}.
The augmentation-based prediction aggregation (APG)
for robustly generating high-quality pseudo labels is described in~\cref{sec:method:subsec:APG}, and the critical retraining strategy (CRS) for adaptively learning from unlabeled data is described in \cref{sec:method:subsec:critical_retraining_strategy}.


\subsection{Preliminary}
\label{Preliminary}
\textbf{Task Definition.}
Given an image sample $x$ in the labeled dataset, its ground truth label $y$ contains information about the category, location, dimension, and orientation of objects visible in $x$.
Semi-supervised Mono3D object detection aims to acquire knowledge from both precisely annotated dataset $\mathcal{D}_l = \{x_i^l, y_i^l\}_{i=1}^{N_l}$ and unlabeled dataset $\mathcal{D}_u = \{x_j^u\}_{j=1}^{N_u}$, where $N_u \gg N_l$.



\textbf{Vanilla Self-Training Scheme.}
As a prominent branch in semi-supervised learning~\cite{self-training-survey, yang2022st++}, self-training works by iteratively optimizing a model with the help of pseudo-labels on the unlabeled samples.
A vanilla self-training~\cite{yang2022st++} pipeline contains three major steps: 1) \textit{Standard Supervised Training} which trains a teacher model $M_t$ on the labeled dataset $\mathcal{D}_l$, 2) \textit{Pseudo Label Generation} which predicts pseudo labels $\{\hat y = M_t(x_j) | x_j \in \mathcal{D}_u\} $ on the unlabeled dataset $\mathcal{D}_u$, and 3) \textit{Retraining with Noisy Labels} which learns a student model $M_s$ for final evaluation.
Using $M_s$ as the new teacher, the step 2 and 3 can be repeated until satisfactory performance is obtained.  


In this paper, we elaborately investigate the pseudo label generation (step 2) and retraining strategy (step 3), which are the most crucial parts of the self-training scheme. For simplicity, we don't iteratively perform step 2 and 3.


\subsection{Augmentation-Based Prediction Aggregation}
\label{sec:method:subsec:APG}

To obtain high-quality pseudo labels, previous 2D semi-supervised learning methods~\cite{wang2021fcos3d,xu2021softteacher,li2022stmono3d, zhou2022denseteacher,chen2022labelmatch} resort to a suitable threshold $\tau{}$ to filter predicted boxes.
However, it is non-trivial to determine an optimal threshold for each different method, especially in the ill-posed Mono3D object detection. A higher threshold may bring tons of false negatives (FN), decreasing the quantity of useful pseudo labels. In contrast, a lower threshold may introduce more false positives (FP), resulting in adverse noises. In order to alleviate the dependency on such a handcrafted threshold, we propose the APG strategy to effectively aggregate predictions from different observations of the same image sample to improve the robustness of pseudo-label generation.

\begin{figure}[t]
    \centering
    \footnotesize
    \includegraphics[width=0.97\linewidth]{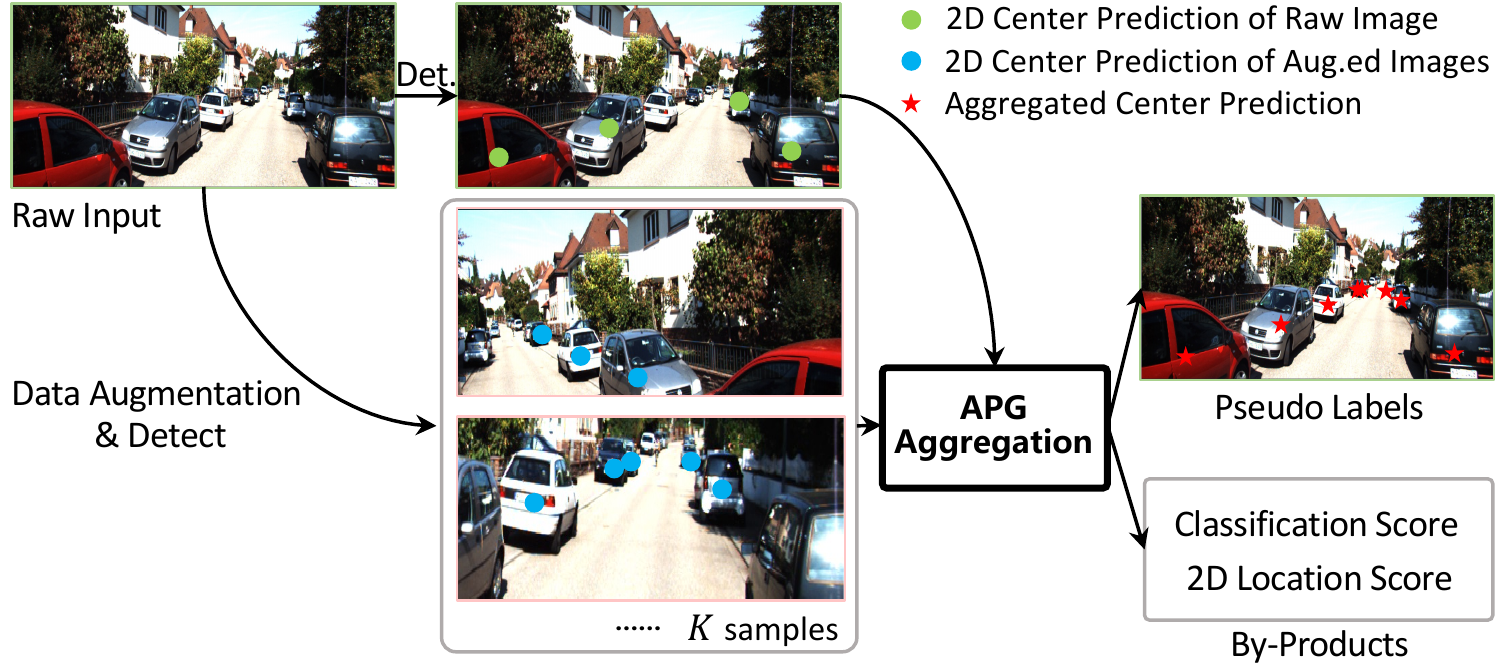}
    \caption{\textbf{Illustration of APG.} We aggregate the predictions from $K$ different transformations of an unlabeled image. The by-product reliability scores estimated by MLE is adopted in retraining to measure the uncertainty of pseudo labels. } 
    \label{fig::method1}
\end{figure}


The proposed algorithm, that is illustrated in Alg.~\ref{alg:st} and Fig.~\ref{fig::method1}, consists of three steps: 

\textbf{1)} Firstly, given an input image from the unlabeled dataset $x_j^u\in \mathcal{D}_u$, the teacher model $M_t$ predicts the detection results for $x_j^u$ and its $K$ augmented images. Let $\mathcal{P}_r$ denote the raw prediction of $x_j^u$, and $\mathcal{P}_f^0$ and $\{\mathcal{P}_f^k\}_{k=1}^{K}$ represent the post-processed (by the pre-defined threshold $\tau$) predictions of $x_j^u$ and the augmented images, respectively. 

\textbf{2)} Secondly, for each prediction $p_i$ in $\mathcal{P}_f^{0}$, we apply the kNN clustering algorithm to find its nearest neighbors in $\{\mathcal{P}_f^k\}_{k=1}^K$, that forms a cluster. $p_i$ is considered as a pseudo label for $x_j^u$. Intuitively, the number of assigned predictions $n$ in the cluster indicates the difficulty degree in detecting an object, and the variance $\sigma$ by Maximum Likelihood Estimation (MLE) measures the uncertainty of $p_i$. With the classification score $s$, these by-products are combined by Eq.~\ref{eq:weight} to demonstrate $p_i$'s reliability, which is then used to weight the loss of each unlabeled data in retraining.
\begin{equation}
    w = \gamma_1 \times s + (1 - \gamma_1) \times \exp{(-\frac{\sigma}{n} * \gamma_2)},
    \label{eq:weight}
\end{equation}
We set $\gamma_1 = 0.6$ and $\gamma_2 = 6$ in our model, respectively.

\textbf{3)} Finally, for the unused predictions in $\{\mathcal{P}_f^k\}_{k=1}^K$, they would be self-clustered. The cluster centers are treated as reference points, whose closest prediction in $\mathcal{P}_r$ are selected as pseudo labels. Their uncertainties are measured by Eq.~\ref{eq:weight}.

Moreover, inspired by successful attempts at auto data augmentations~\cite{cubuk2018autoaugment,lim2019fastaugment}, we resort to the
Tree-Structured Parzen Estimators (TPE)~\cite{bergstra2011algorithms} to automatically pick the $K$ transformations and their hyper-parameters (\emph{e.g.}, resize ratio). More details are presented in supplementary materials.





\setlength{\textfloatsep}{0.2cm}
\begin{algorithm}[t]
\linespread{1.0}\selectfont
\caption{\label{alg:st}APG Aggregation Pseudocode}

\SetAlgoNoLine

\KwIn{Predictions of different observations,\\
Raw prediction $\mathcal{P}_r$ of an unlabeled image,

Filtered prediction $\mathcal{P}_f^0$, $N = \text{len}(\mathcal{P}_f^0)$, 

Filtered predictions $\{\mathcal{P}_f^k\}_{k=1}^K$ of augmented images,

Threshold $\tau$ for kNN
}

\vspace{0.05cm}
\KwOut{Aggregated prediction $\mathcal{P}$}
\vspace{0.1cm}
Initialize set $\mathcal{S} \leftarrow \{\{p_1\},\cdots, \{p_N\}\},p_n\in \mathcal{P}_f^0$


\For {image observation $k\in \{1, \dots, K\}$} {
    \For {prediction $p_i\in\mathcal{P}_f^k$} {
    $(\text{index } j, \text{distance } l)$ $\leftarrow$ kNN$(p_i, \mathcal{S})$
    
        \eIf{$l < \tau$}{
        $\mathcal{S}_j \leftarrow \mathcal{S}_j \cup \{p_i\}$ \# append $p_i$ to clusters
        }{
        $\mathcal{S}_j \leftarrow \mathcal{S}_j \cup \{\{p_i\}\}$ \# create new clusters
        }
    }
}

\For {loop index n, cluster set $\{p^m\}_{m=1}^M \in \mathcal{S}$ } {
location $\mu$, variance $\sigma = $ MLE $(\{p^m\}_{j=m}^M)$

\eIf{$n < N$}{
    $ \mathcal{P} \leftarrow \mathcal{P} \cup \{(\mathcal{S}_n[0], \sigma)\}$
}
{
    $\mathcal{P} \leftarrow \mathcal{P} \cup \{(\text{NearestSearch}(\mu, \mathcal{P}_r), \sigma)\}$
}
}

\Return {$\mathcal{P}$}
\end{algorithm}

\subsection{Critical Retraining Strategy}
\label{sec:method:subsec:critical_retraining_strategy}

Generated pseudo labels inevitably contain noises, thus it is crucial to find informative ones that benefit model evolution. Previous methods (\eg,~\cite{xu2021softteacher}) use the box jitter scores as proxies for the pseudo label quality. However, it might be hard to replicate the success of such strategy in the Mono 3D detection due to the inferior performance of the teacher model $M_t$.
The uncertainty measurements of pseudo labels provided by the APG module can enhance the stability of student model retraining, but it still suffers from the fixed weight of each sample. We argue that the contribution of each sample during model training should adapt to the model's state as training proceeds~\cite{crs_ref1, crs_ref2}.

To this end, we propose a learning-based critical module to adaptively find the informative unlabeled data, which may provide a new perspective for semi-supervised Mono3D object detection to retrain a better student model. 


\begin{figure}[!t]
    \centering
    \footnotesize
    \includegraphics[width=0.98\linewidth]{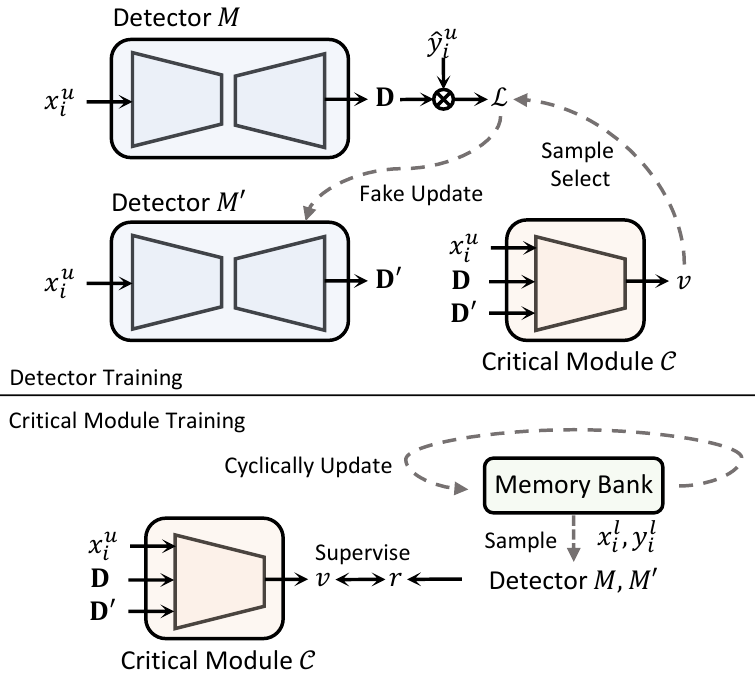}
    \caption{\textbf{Illustration of CRS.} We adopt a critical module to discriminate whether a sample from the unlabeled data benefits model convergence. The memory bank is cyclically updated.} 
    \label{fig::method2}
\end{figure}


Specifically, the critical module first evaluates the effect of a training sample from the unlabeled dataset, and then assigns it with a 0-1 binary flag indicating whether to back-propagate its gradients. From a reinforcement learning perspective, we regard the Mono3D detector (student model) as an \textit{agent}, the model's weight parameters as the~\textit{state}, the input image and the output of the model as an \textit{observation}. 
At state $\mathcal{S}$, a detection loss $\mathcal{L}_{unsup}$ for the agent can be calculated based on the given observation $\mathcal{O}$.
If the gradients of $\mathcal{L}_{unsup}$ are back-propagated, the state will be updated to $\mathcal{S}^{'}$ and the model output will be updated to $\mathcal{O}^{'}$. The critical module then evaluates whether $\mathcal{S}^{'}$ is the optimal choice of updated $\mathcal{S}$ based on the observations $\mathcal{O}$ and $\mathcal{O}^{'}$.




At each training step, an input image $x_i^u$ from the unlabeled data is fed into the detector $M$ (agent), obtaining the detection predictions $\mathbf{D}$ (classification and regression response maps),
\begin{equation}
    \mathbf{D} = M(x_i^u | \mathcal{S}),
\end{equation}
With the pseudo label $\hat y_i^u$, we can get the training loss,
\begin{equation}
\mathcal{L}_{unsup} = \mathcal{L} (\mathbf{D}, \hat y_i^u),
\end{equation}
We take one `\textit{trial}' gradient descent step to obtain the updated model $M'$ with parameters $\mathcal{S}^{'}$. Then, the critical module evaluates the effectiveness of this update ($\mathcal{S} \rightarrow \mathcal{S}^{'}$),
\begin{equation}
    v = \mathcal{C}(x_i^u, \mathbf{D}, \mathbf{D}^{'} | \Psi),
    \label{eq:cr}
\end{equation}
where $\mathbf{D}^{'}$ is the detection predictions of the updated model $M'$ on $x_i^u$, and $\Psi$ is the parameter of the critical module. During training, we chop off a certain number of samples with the lowest evaluation value $v$ (see Fig.~\ref{fig::method2}).


To guarantee the critical module can provide reliable feedback, we propose a reward function to supervise the training of the critical network,
\begin{equation}
    r = \mathcal{L}(M(x_i| \mathcal{S}), y_i) - \mathcal{L}(M'(x_i|\mathcal{S}^{'}), y_i),
    \label{eq:reward}
\end{equation}
where $(x_i, y_i)$ denotes samples from the training set of the labeled dataset $\mathcal{D}_l$. The L2 loss~\cite{l2loss} is applied to $v$ and $r$ for supervising the learning of critical module. During training, we alternately update the detector and critical module. 

Notably, it's impractical to evaluate all samples to get a reliable reward $r$ due to the unaffordable computation cost. Motivated by the self-supervised method MoCo~\cite{chen2020mocov2}, we employ a  memory bank (queue) to buffer the training samples in $\mathcal{D}_l$ and cyclically update it. After tons of steps updating, the knowledge of all samples for evaluation are encoded to the weight parameters of the critical network, making it capable of predicting accurate indicator.


\section{Experiments}

In this section, we first recap the experimental setup in Sec.~\ref{exp:setup}. Then, we respectively present the evaluation results (Sec.~\ref{exp:main_res}), ablation studies and analysis (Sec.~\ref{exp:abl_dis}) to demonstrate the effectiveness of the proposed methods.

\begin{table*}[t!]
\centering
\vspace{0.15cm}
\scalebox{0.85}{%
\begin{tabular}{l|c|c|ccc|ccc|ccc}
\hline
\multirow{2}{*}{Method} & \multirow{2}{*}{Reference} & \multirow{2}{*}{Extra data} & \multicolumn{3}{c|}{Test $AP_{3D}$} & \multicolumn{3}{c|}{Test $AP_{BEV}$} & \multicolumn{3}{c}{Val $AP_{3D}$} \\ 
& & & Easy & Mod. & Hard & Easy & Mod. & Hard & Easy & Mod. & Hard \\
\hline
PatchNet~\cite{ma2020patchnet} & ECCV2020 & \multirow{3}{*}{Depth}   & 15.68 & 11.12 & 10.17 & 22.97 & 16.86 & 14.97 & -  & -  & - \\ 
D4LCN~\cite{ding2020d4lcn} & CVPRW2020 &             & 16.65 & 11.72 & 9.51  & 22.51 & 16.02 & 12.55 & - & - & - \\
DDMP-3D~\cite{wang2021ddmp} & CVPR2021 &                     & 19.71 & 12.78 & 9.80  & 28.08 & 17.89 & 13.44 & - & - & - \\
\hline
Kinematic3D~\cite{brazil2020kinematic} & ECCV2020 & Multi-frames      & 19.07 & 12.72 & 9.17  & 26.69 & 17.52 & 13.10 & 19.76 & 14.10 & 10.47 \\
\hline
MonoRUn~\cite{chen2021monorun} & CVPR2021 & \multirow{3}{*}{LiDAR}    & 19.65 & 12.30 & 10.58 & 27.94 & 17.34 & 15.24 & 20.02 & 14.65 & 12.61 \\
CaDDN~\cite{reading2021caddn} & CVPR2021 &   & 19.17 & 13.41 & 11.46 & 27.94 & 18.91 & 17.19 & 23.57 & 16.31 & 13.84 \\
MonoDTR~\cite{huang2022monodtr}& CVPR2022 &     & 21.99 & 15.39 & 12.73 & 28.59 & 20.38 & 17.14 & 24.52 & 18.57 & 15.51\\
\hline
AutoShape~\cite{liu2021autoshape}& ICCV2021 &  CAD  & 22.47 & 14.17 & 11.36 & 30.66 & 20.08 & 15.59 & 20.09 & 14.65 & 12.07 \\
\hline
SMOKE~\cite{liu2020smoke}& CVPRW2020 &     \multirow{8}{*}{None}                           & 14.03 & 9.76  & 7.84  & 20.83 & 14.49 & 12.75 & 14.76 & 12.85 & 11.50 \\
MonoPair~\cite{chen2020monopair}& CVPR2020 &                          & 13.04 & 9.99  & 8.65  & 19.28 & 14.83 & 12.89 & 16.28 & 12.30 & 10.42\\
RTM3D~\cite{li2020rtm3d}& ECCV2020 &         & 13.61 & 10.09 & 8.18  & -     & -     & -     & 19.47 & 16.29 & 15.57 \\
PGD~\cite{wang2022pgd}& CoRL2022 &                          & 19.05 & 11.76 & 9.39 & 26.89 & 16.51 & 13.49 & 19.27 & 13.23 & 10.65 \\ 
MonoRCNN~\cite{shi2021monorcnn}& ICCV2021 &                          & 18.36 & 12.65 & 10.03 & 25.48 & 18.11 & 14.10 & 16.61 & 13.19 & 10.65 \\ 
Zhang \textit{et al.}$_{\text{DLE}}$~\cite{zhang2022dimension}& CVPR2022 &                            & 20.25 & 14.14 & 12.42 & 28.85 & 17.72 & 17.81 & 20.82 & 15.64 & 13.82 \\ 
GUPNet~\cite{lu2021gup}& ICCV2021 &                              & 20.11 & 14.20 & 11.77 & -     & -     & -     & 22.76 & 16.46 & 13.72 \\ 
HomoLoss$_{\text{FLEX}}$~\cite{gu2022homography}& CVPR2022 &                            & 21.75 & 14.94 & 13.07 & 29.60 & 20.68 & 17.81 & 23.04 & 16.89 & 14.90 \\
\hline
Mix-Teaching$_{\text{FLEX}}$~\cite{yang2022mix}\dag & Arxiv2022 &           Unlabeled    & 21.88 & 14.34 & 11.86 & 30.52 & 19.51 & 16.45 & 25.54 & 19.35 & 16.51 \\ 

\hline
\textcolor{brown}{MonoDLE}~\cite{ma2021monodle}& CVPR2021 & None                           & 17.23 & 12.26 & 10.29 & 24.79 & 18.89 & 16.00 & 17.45 & 13.66 & 11.68 \\ 
\cellcolor{lightgray!15}{\textbf{3DSeMo$_{\text{DLE}}$}}& \cellcolor{lightgray!15}{Ours} & \cellcolor{lightgray!15}{Unlabeled} & \cellcolor{lightgray!15}{\textbf{23.11}} & \cellcolor{lightgray!15}{\textbf{15.58}} & \cellcolor{lightgray!15}{\textbf{13.58}} & \cellcolor{lightgray!15}{\textbf{30.99}} & \cellcolor{lightgray!15}{\textbf{21.78}} & \cellcolor{lightgray!15}{\textbf{18.64}} & \cellcolor{lightgray!15}{\textbf{22.87}} & \cellcolor{lightgray!15}{\textbf{17.65}} & \cellcolor{lightgray!15}{\textbf{14.83}} \\
\cellcolor{lightgray!40}{\textit{Improvement}} & \cellcolor{lightgray!40}{\textit{Ours}} & \cellcolor{lightgray!40}{\textit{v.s. baseline}} &\cellcolor{lightgray!40}{\textcolor[RGB]{61,145,64}{+5.88}} &\cellcolor{lightgray!40}{\textcolor[RGB]{61,145,64}{+3.32}} &\cellcolor{lightgray!40}{\textcolor[RGB]{61,145,64}{+3.29}} &\cellcolor{lightgray!40}{\textcolor[RGB]{61,145,64}{+6.20}} &\cellcolor{lightgray!40}{\textcolor[RGB]{61,145,64}{+2.89}} &\cellcolor{lightgray!40}{\textcolor[RGB]{61,145,64}{+2.64}} &\cellcolor{lightgray!40}{\textcolor[RGB]{61,145,64}{+5.42}} &\cellcolor{lightgray!40}{\textcolor[RGB]{61,145,64}{+3.99}} &\cellcolor{lightgray!40}{\textcolor[RGB]{61,145,64}{+3.15}}\\
\textcolor{brown}{MonoFlex}~\cite{zhang2021monoflex}\dag& CVPR2021 &  None                        & 19.94 & 13.89 & 12.07 & 28.23 & 19.75 & 16.89 & 23.44 & 17.43 & 14.67 \\ 
\cellcolor{lightgray!15}{\textbf{3DSeMo$_{\text{FLEX}}$}} & \cellcolor{lightgray!15}{Ours} & \cellcolor{lightgray!15}{Unlabeled} & \cellcolor{lightgray!15}{\textbf{23.55}} & \cellcolor{lightgray!15}{\textbf{15.25}} & \cellcolor{lightgray!15}{\textbf{13.24}} & \cellcolor{lightgray!15}{\textbf{32.57}} & \cellcolor{lightgray!15}{\textbf{21.21}} & \cellcolor{lightgray!15}{\textbf{18.07}} & \cellcolor{lightgray!15}{\textbf{27.35}} & \cellcolor{lightgray!15}{\textbf{20.87}} & \cellcolor{lightgray!15}{\textbf{17.66}} \\
\cellcolor{lightgray!40}{\textit{Improvement}}& \cellcolor{lightgray!40}{\textit{Ours}} & \cellcolor{lightgray!40}{\textit{v.s. baseline}} &\cellcolor{lightgray!40}{\textcolor[RGB]{61,145,64}{+3.61}} &\cellcolor{lightgray!40}{\textcolor[RGB]{61,145,64}{+1.36}} &\cellcolor{lightgray!40}{\textcolor[RGB]{61,145,64}{+1.17}} &\cellcolor{lightgray!40}{\textcolor[RGB]{61,145,64}{+4.34}} &\cellcolor{lightgray!40}{\textcolor[RGB]{61,145,64}{+1.46}} &\cellcolor{lightgray!40}{\textcolor[RGB]{61,145,64}{+1.18}} &\cellcolor{lightgray!40}{\textcolor[RGB]{61,145,64}{+3.91}} &\cellcolor{lightgray!40}{\textcolor[RGB]{61,145,64}{+3.44}} &\cellcolor{lightgray!40}{\textcolor[RGB]{61,145,64}{+2.99}}\\
\hline
\end{tabular}
}
\vspace{0.2cm}
\caption{\textbf{Comparision with state-of-the-art (SOTA) Methods.} We present the evaluation results of `CAR' in the KITTI test and validation sets. \dag denotes the reproduction results. For fair comparisons, we train Mix-Teaching with the same data volume as our method.}.
\label{tab:kitti_test}
\vspace{-1em}
\end{table*}





\subsection{Experimental Setup}
\label{exp:setup}
\textbf{Dataset.} We mainly conduct our experiments on KITTI~\cite{geiger2012kitti}, which contains 7,481 images for training and 7,518 images for testing. Following \cite{chen20153d}, we split the original training set into 3,712 training and 3,769 validation samples.
We picked 151 unlabeled video sequences from KITTI. After removing the duplicated samples in the training set, we obtained 33,507 unlabeled samples for semi-supervised training purposes, roughly 10 times larger than the annotated training set.
A smaller subset of the training data is held out for the TPE hyper-parameter search. It is observed that previous works on semi-supervised Mono 3D detection only evaluate KITTI~\cite{geiger2012kitti}. To further show the generality of our method, we design a toy experiment on nuScenes (see Sec.~\ref{sec:nus_exp}).



\textbf{Metrics.} In our experiments, we utilize the average precision (AP) as the metrics (both in 3D and bird’s eye view) to evaluate and compare different methods. To prove the effectiveness of the proposed APG, we calculate the detection recall to measure the qualities of the generated pseudo labels (see Sec.~\ref{sec:exp:subsec:APG}). Following \cite{simonelli2019disentangling}, all evaluation results on validation and test sets are based on AP@40.



\textbf{Implementation Details.} We integrate the proposed semi-supervised framework to classical Mono3D detectors MonoDLE~\cite{ma2021monodle} and MonoFlex~\cite{zhang2021monoflex}. Unless otherwise specified, the proposed APG augments an input image from the unlabeled dataset to $K=9$ different views. While the initial threshold for filtering detection boxes is set as 0.65, other predictions with confidence scores lower than 0.65 will be used in the center aggregation algorithm (see Sec.~\ref{sec:method:subsec:APG}). For the proposed CRS, we construct the critical module with ResNet-18~\cite{he2016resnet} network and modify the output dimension of the last fully connected layer to 1. Other layers are initialized with the standard pre-trained weights on ImageNet~\cite{deng2009imagenet}. Notably, the critical module is not used during inference. For a batch size of 8, we chop off the 2 samples with lowest evaluation value $v$ in CRS training. For retraining, we initialize the student model with the weights from the teacher model trained on the labeled dataset. 


For fair comparisons, we reproduce the baseline methods MonoDLE~\cite{ma2021monodle} and MonoFlex~\cite{zhang2021monoflex} based on the official codes provided by the authors. While most Mono3D methods are trained on a single GPU, we adopt 8 A6000 GPUs in all experiments to facilitate training with a larger data volume. 
Ablations are conducted based on MonoDLE unless otherwise specified. Configs will be released. 



\subsection{Main Results}
\label{exp:main_res}

\begin{table}[t]
    \centering
    \scalebox{0.8}{%
    \begin{tabular}{c|ccc|ccc}
        \hline
        \multirow{2}{*}{Method} & \multicolumn{3}{c|}{Ped. $\text{AP}_{3D}$ $\text{IoU}\geqslant0.5$} & \multicolumn{3}{c}{Cyc. $\text{AP}_{3D}$ $\text{IoU}\geqslant0.5$} \\
        & Easy & Mod. & Hard & Easy & Mod. & Hard \\ \hline
        \textcolor{brown}{Baseline} &  9.64 &	6.55 &	5.44  & 4.59 & 2.66 & 2.45 \\
        \cellcolor{lightgray!15}{\textbf{Ours}}  & \cellcolor{lightgray!15}{\textbf{10.78}} & \cellcolor{lightgray!15}{\textbf{7.26}} & \cellcolor{lightgray!15}{\textbf{6.05}} & \cellcolor{lightgray!15}{\textbf{7.04}} & \cellcolor{lightgray!15}{\textbf{4.24}} & \cellcolor{lightgray!15}{\textbf{3.56}} \\
        \cellcolor{lightgray!40}{\textit{Improvement}}  &\cellcolor{lightgray!40}{\textcolor[RGB]{61,145,64}{+1.14}} &\cellcolor{lightgray!40}{\textcolor[RGB]{61,145,64}{+0.71}} &\cellcolor{lightgray!40}{\textcolor[RGB]{61,145,64}{+0.61}} &\cellcolor{lightgray!40}{\textcolor[RGB]{61,145,64}{+2.45}} &\cellcolor{lightgray!40}{\textcolor[RGB]{61,145,64}{+1.58}}&\cellcolor{lightgray!40}{\textcolor[RGB]{61,145,64}{+1.11}}\\

        \hline
    \end{tabular}}
    \vspace{0.2cm}
    \caption{`Pedestrian' and `Cyclist' Performances in KITTI test set.}
\label{tab:kitti_ped_cyc}
\end{table}

We apply the proposed framework to MonoDLE~\cite{ma2021monodle} and MonoFlex~\cite{zhang2021monoflex}, and evaluate our methods on the official test and validation sets of KITTI.  Tab.~\ref{tab:kitti_test} and Tab.~\ref{tab:kitti_ped_cyc} present quantitative comparisons of our method with other state-of-the-art counterparts on the KITTI leaderboard. It shows that by effectively leveraging larger volumes of unlabeled data, our proposed semi-supervised strategy significantly boosts the performance of the baseline methods. In particular, our approach respectively improves the baseline MonoDLE $\mathtt{AP_{3D} (Mod.)}$ and $\mathtt{AP_{BEV} (Mod.)}$ by +3.32\%/+2.89\% on the test set without any tricks (\emph{e.g.,} test-time augmentation). The gains on $\mathtt{AP (Easy)}$ of our 3DSeMo$_{\text{DLE}}$ surprisingly exceeds +5\% on all metrics and data splits. When integrating our method to MonoFlex~\cite{liu2022monocon}, it achieves gains of +3.61/1.36 on $\mathtt{AP_{3D} (Easy/Mod.)}$, respectively, evidencing the generality of our framework. 








\subsection{Ablation Studies and Analysis}
\label{exp:abl_dis}

This section presents more in-depth analyses to demonstrate the effectiveness of each proposed component.

\subsubsection{Influence of the Unlabeled Data Volume}
We compare the fully supervised baseline with our semi-supervised model trained with different amounts of unlabeled data in Tab.~\ref{tab:label_unlabel_ratio}. It shows that our semi-supervised method outperforms the baseline under different unlabeled data volume, which proves that our method can effectively exploit the useful information in unlabeled data and the generated pseudo labels. Notably, even only using 20\% unlabeled data (about 3350 samples), our approach can obtain +1.55\% $\mathtt{AP_{3D} (Mod.)}$ and +1.25\% $\mathtt{AP_{BEV} (Mod.)}$ improvements on car and pedestrian, respectively. When using all the unlabeled data, our method outperforms the baseline with a favorable margin, demonstrating the profound potentials of semi-supervised learning in boosting Mono3D detection.

\begin{table}[t]
\renewcommand{\tabcolsep}{3.6mm}
    \centering
    \scalebox{0.88}{%
    \begin{tabular}{c|ccc}
        \hline
        Method & Car & Ped. & Cyc.\\ \hline
        Baseline &  13.66 & 4.45 & 2.52 \\
        + 20\% Unlabeled Data & 15.21 & 5.21 & 2.51 \\
        + 50\% Unlabeled Data & 16.50 & 6.37 & 2.76 \\
        + 100\% Unlabeled Data & \textbf{17.65} & \textbf{8.25} & \textbf{3.47} \\
        \hline
    \end{tabular}}
    \vspace{0.2cm}
    \caption{Influence of the Unlabeled Data Volume (AP$_{3D}$ Mod.).}
\label{tab:label_unlabel_ratio}
\end{table}


\subsubsection{Component-wise Analysis}
To understand the effect of each component, we incrementally apply the proposed APG and CRS to the baseline detector MonoDLE~\cite{ma2021monodle} with $\mathtt{Car~ AP_{3D} (Mod.)}$ of 13.66. As shown in Tab.~\ref{tab:warmup}, the vanilla self-training strategy (see Sec.~\ref{Preliminary}) improves the baseline model for 2.11\% on $\mathtt{Car~ AP_{3D} (Mod.)}$ without bells and whistles. Subsequently, we apply the proposed APG and CRS to the model, respectively. It shows that both of them can significantly improve $\mathtt{Car~ AP_{3D} (Mod.)}$ by about 2.5\%, which validates our argument that robust pseudo label generation and finding informative samples are both crucial for semi-supervised Mono3D object detection. Last but not least, while maintaining the superiority on $\mathtt{Car}$, combining the proposed APG and CRS can pre-eminently improve  $\mathtt{Ped. (Mod.)~ AP_{3D}}$ for about 1.4\%. In Mono3D object detection, the pedestrian is more challenging than car because of the much smaller object size, for which a slight prediction shift leads to drastic degradation of IoU. The pseudo labels of pedestrian thus contain much more noise. Therefore, the gains on pedestrian prove CRS's ability in adaptively selecting informative samples from severely noisy pseudo labels.

\begin{table}[t]
    \centering
    \scalebox{0.73}{%
    \begin{tabular}{c|ccc|ccc}
        \hline
        \multirow{2}{*}{Method} & \multicolumn{3}{c|}{Car $\text{AP}_{3D}$ $\text{IoU}\geqslant0.7$} & \multicolumn{3}{c}{Ped. $\text{AP}_{3D}$ $\text{IoU}\geqslant0.5$} \\
        & Easy & Mod. & Hard & Easy & Mod. & Hard \\ \hline
        Baseline &  16.77 & 13.66 & 11.54 & 5.53 & 4.45 & 3.33 \\
        Plainest Self-Training & 20.14 & 15.77 & 13.27 & 7.27 & 5.99 & 4.74 \\
        
        + CRS & 22.64 & 17.53 & 14.59 & 9.43 & 6.81 & 5.71\\
        + APG & 22.71 & 17.56 & 14.68 & 9.35 & 6.77 & 5.58\\
        + APG + CRS & \textbf{22.87} & \textbf{17.65} & \textbf{14.83} & \textbf{10.99} & \textbf{8.25} & \textbf{6.72} \\
        \hline
    \end{tabular}}
    \vspace{0.2cm}
    \caption{\textbf{Component-wise Analysis.} We compare the effects of each proposed component to demonstrate the effectiveness and rationality of our framework.}
\label{tab:warmup}
\end{table}




\begin{figure}[t]
    \centering
    \footnotesize
    \includegraphics[width=0.98\linewidth]{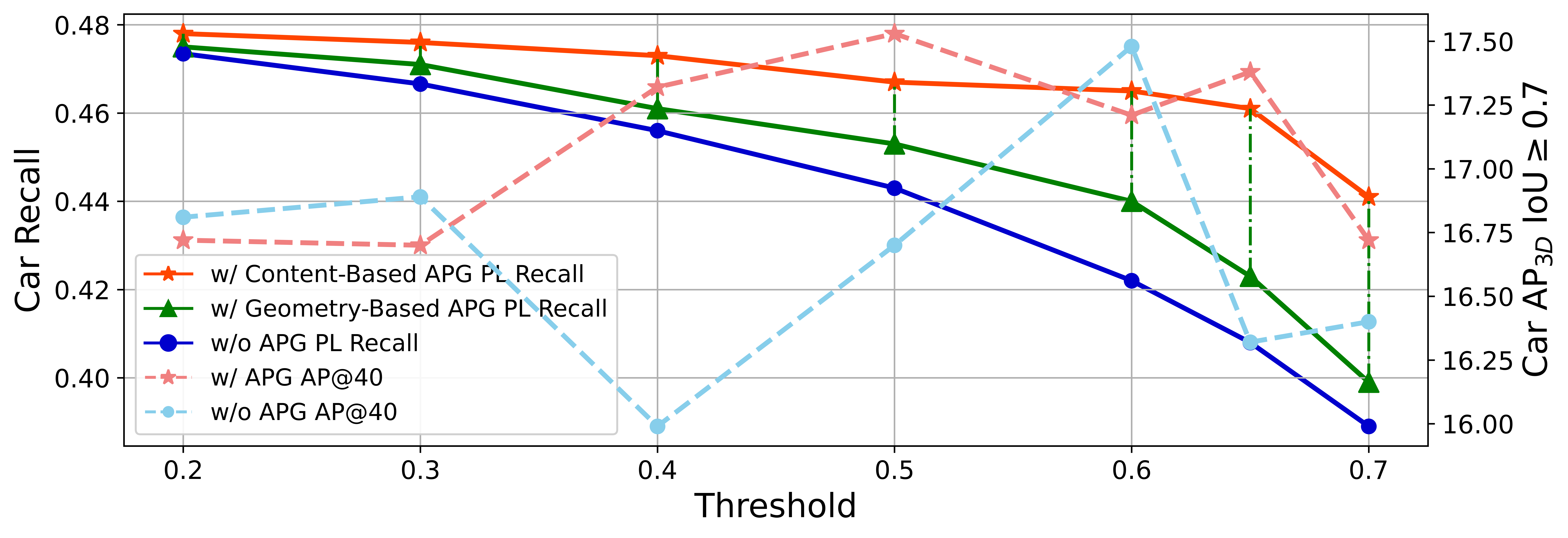}
    \caption{\textbf{Effectiveness of APG.} We demonstrate the robustness of APG on both pseudo label quality (recall) and 3D object detection performance (AP@40). Results show that our method is less sensitive to threshold change.}
    \label{fig:APG}
\end{figure}

\subsubsection{Analysis of APG}
\label{sec:exp:subsec:APG}
\textbf{Robustness.} Previous semi-supervised methods usually filter detection boxes to generate pseudo labels by applying a threshold $\tau$ on the classification score. However, as presented in Fig.~\ref{fig:APG} (the \textcolor{blue}{blue} solid line), it suffers from a drastic degradation on $\mathtt{Recall}$ when enlarging the threshold. Besides, its detection performance is sensitive to threshold change (the \textcolor{cyan}{
cyan} dotted line). In contrast, the performances of our APG (the \textcolor{red}{red} dotted lines) are more stable, which proves its robustness in generating pseudo labels. We select $\tau=0.65$ in our model based on the observation of this experiment, with which the APG can boost the $\mathtt{Car~(Mod.)~ AP_{3D}}$ for 1.06\%, as shown in Tab.~\ref{tab:2dscore}. Though we need to set an initial threshold in APG, our experiment (the \textcolor{red}{red} solid line) show that it is less sensitive to threshold change. 
Fig.~\ref{fig:APG} also proves that geometry-based augmentation (the \textcolor[RGB]{61,145,64}{green} solid line, \textit{i.e.,} geometry transform) is superior to the content-based counterpart (the \textcolor{red}{red} solid line, \textit{i.e.,} color enhencement) in improving the quality of the generated pseudo labels.
This may attribute to their different mechanism that content-based transformations only marginally modify the context, while geometry-based transformations can significantly migrate the position and scale distribution of objects, which are the common reasons for false negatives in Mono3D object detection (see Fig.~\ref{fig::teaser} again). Notably, the transformations are automatically learned by TPE, which can be effortlessly integrated into other detectors. All details, codes, and results about TPE will be released for reproduction.


\begin{table}[t]
    \centering
    \scalebox{0.74}{%
    \begin{tabular}{c|cc|ccc|ccc}
        \hline
        & \multicolumn{2}{c|}{Method} & \multicolumn{3}{c|}{Car $\text{AP}_{3D}$ $\text{IoU}\geqslant0.7$} & \multicolumn{3}{c}{Ped. $\text{AP}_{3D}$ $\text{IoU}\geqslant0.5$} \\
        &+cls. & +loc.  & Easy & Mod. & Hard & Easy & Mod. & Hard \\ \hline
        \ding{172}&\multicolumn{2}{c|}{w/o APG} & 21.75 & 16.32 & 14.15 & 8.34 & 6.04 & 4.80 \\ \hline
        \ding{173}&\multicolumn{2}{c|}{w/ APG}  &  22.66 & 17.38 & 14.67 & 7.71 & 5.88 & 4.74 \\
        \ding{174}&\ding{51}&  & 22.51 & 17.44 & 14.63 & 9.02 & 6.52 & 5.33 \\
        \ding{175}&\ding{51}&  \ding{51}& \textbf{22.71} & \textbf{17.56} & \textbf{14.68} & \textbf{9.35} & \textbf{6.77} & \textbf{5.58} \\
        \hline
        
        \ding{176}&\multicolumn{2}{c|}{DenseTeacher~\cite{zhou2022denseteacher}}& 22.83 & 17.12 & 14.36 & 7.78 & 5.94 & 4.75 \\
        \hline
    \end{tabular}}
    \vspace{0.2cm}
    \caption{\textbf{Effectiveness of Reweighting Strategy.} With the by-product of APG to reweight samples during retraining, the student model obtains impressive gains, especially on pedestrian.}
\label{tab:2dscore}
\end{table}


\noindent\textbf{Influence of sample weight.} While APG improves the overall recall of pseudo labels, it inevitably introduces more noise to challenging categories (\textit{e.g.,} pedestrian) as shown in Tab.~\ref{tab:2dscore}. As a result, the $\mathtt{Ped.~(Mod.)~ AP_{3D}}$ drops 0.16\% (\ding{173} $v.s.$ \ding{172}). To alleviate this, we weight the loss of each unlabeled sample in the retraining phase with the by-product clues from the proposed APG (see Sec.~\ref{sec:method:subsec:APG} and Eq.~\ref{eq:weight}). As shown in Tab.~\ref{tab:2dscore}, when adopting classification score to weight samples as in previous works~\cite{xu2021softteacher}, it can improve the performance of hard category ( \ding{174} $v.s.$ \ding{173}). Introducing the clues from APG can further enhance the detection of pedestrian, obtaining 0.89\% improvement on the $\mathtt{Ped.~(Mod.)~ AP_{3D}}$ (\ding{175} $v.s.$ \ding{173}).
Threshold-free approaches, such as DenseTeacher~\cite{zhou2022denseteacher}, however, does not bring benefits to the Mono3D SSL task. The discrepancies between 2D and 3D detections is to blame for such failure.


\subsubsection{Analysis of CRS}
\label{sec:exp:subsec:crs}
\begin{figure}[t]
    \centering
    \footnotesize
    \includegraphics[width=0.98\linewidth]{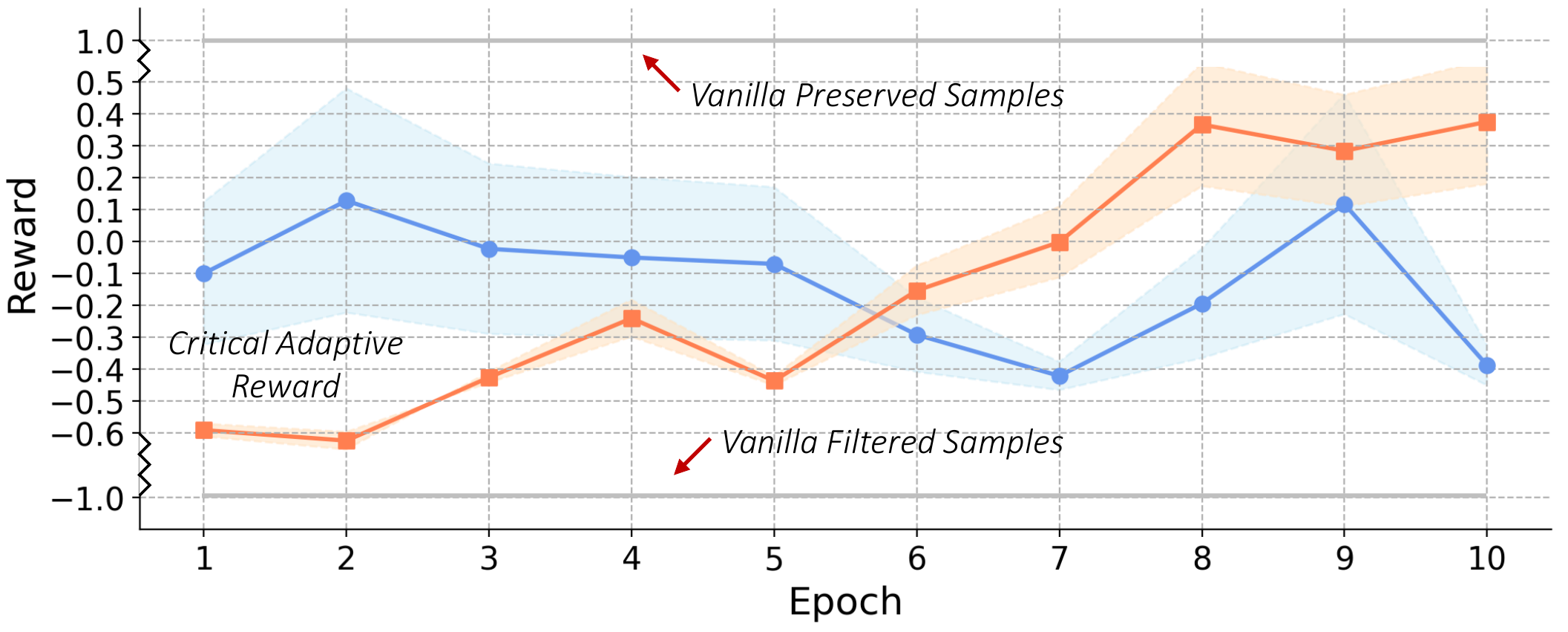}
    \caption{\textbf{Adaptive Reward of CRS.} Vanilla filter-based strategies invariably drop out or preserve a samples whereas our proposed critical module predicts adaptive rewards during retraining.}
    \label{fig:reward}
\end{figure}

\textbf{Different strategies for selecting informative samples.} The proposed CRS aims to adaptively separate informative samples from noisy ones. To demonstrate the superiority of CRS, we compare against some alternative strategies which have demonstrated success in other tasks. The compared counterparts include 1) filtering samples with the quality score of pseudo labels introduced in Eq.~\ref{eq:weight}, 2) the bbox jitter proposed for 2D detection in \cite{xu2021softteacher}. We tailor the 2D box jitter strategy for 3D detection, and details are presented in the supplementary material. As shown in Tab.~\ref{tab:crs}, bbox jitter causes performance degradation because of its unreliable quality measurement for pseudo labels (\ding{173}  \textit{v.s.} \ding{172}). \ding{174} throws away unreality samples based on classification and location scores (see Eq.~\ref{eq:weight}). It shows that \ding{174} only slightly improves pedestrian detection performance, however still lagging behind our proposed CRS (\ding{176}). Besides the unreliability of detection scores and box jitter in Mono3D, another underlying reason for the advance of CRS is that \ding{172} and \ding{173} are static strategies where the filtering indicator of a sample holds along the retraining phase. Conversely, the indicator learned by the critical module changes in different retraining timestamps, as shown in Fig.~\ref{fig:reward}. It both intuitively and theoretically makes sense that the importance of a sample should be mutative in training. 




\begin{table}[t]
    \centering
    \begin{adjustbox}{width=0.99\linewidth,center}
    \begin{tabular}{@{}c|c|ccc|ccc@{}}
        \hline
        & \multirow{2}{*}{Method} & \multicolumn{3}{c|}{Car $\text{AP}_{3D}$ $\text{IoU}\geqslant0.7$} & \multicolumn{3}{c}{Ped. $\text{AP}_{3D}$ $\text{IoU}\geqslant0.5$} \\
        & & Easy & Mod. & Hard & Easy & Mod. & Hard \\ \hline
        \ding{172} & Baseline & 22.71 & 17.56 & 14.68 & 9.35 & 6.77 & 5.58 \\
        \ding{173} & + 3D bbox jitter filter ~\cite{xu2021softteacher}& 22.50 & 16.71 & 14.40 & 8.36 & 6.30 & 5.07 \\
        \ding{174} & + cls.\&loc. weight filter & 21.86 & 17.20 & 14.31 & 9.04 & 7.25 & 5.75 \\
        \ding{175} & + CRS filter w/o critical module & 21.41 & 16.58 & 14.05 & 8.05 & 6.23 & 5.01 \\
        \ding{176} & + CRS filter w/ critical module& \textbf{22.87} & \textbf{17.65} & \textbf{14.83} & \textbf{10.99} & \textbf{8.25} & \textbf{6.72} \\
        \hline
    \end{tabular}
    \end{adjustbox}
    \vspace{0.2cm}
    \caption{\textbf{Effectiveness of CRS.} We compare our critical module with other counterparts in filtering pseudo labels. The prominent improvement on pedestrian indicates that the CRS can effectively find informative samples during training.}
\label{tab:crs}
\end{table}

\noindent\textbf{Learnable or not.} The proposed CRS learns the filtering indicator with a learnable critical module (Eq.~\ref{eq:cr}). Yet intuitively, we can simply determine the contribution of a sample by the training loss before and after the model updating with Eq.~\ref{eq:reward}.  To validate the necessity of the proposed scheme, we prohibit the critical module and directly leverage the reward calculated in Eq.~\ref{eq:reward} as the indicator to select samples during retraining. As shown in Tab.~\ref{tab:crs} \ding{175}, unsurprisingly, this naive strategy degrades the overall performance because of its biased optimization objective. In particular, the strategy of \ding{175} can only access several samples during calculating the reward, lacking the global vision of the evaluation set. In contrast, the learning-based critical module encodes the knowledge of the whole dataset to its weights parameters through cyclically updating the memory bank, which
 can provide stable and effective filtering indicators for model retraining (see \ding{176}). 

\begin{table}[t]
\renewcommand{\tabcolsep}{3.2mm}
    \centering
    \scalebox{0.88}{%
    \begin{tabular}{c|ccc|cc}
        \hline
        \multirow{2}{*}{Method} & \multicolumn{5}{c}{KITTI Validation}  \\ \cline{2-6}
        & 0-20 & 20-40 & 40-$\infty$ & Avg. & AP$_{3D}$\\ \hline 
        Baseline &  0.511 & 1.243 & 2.639 & 1.172 & 13.66 \\
        \textbf{Ours} & \textbf{0.446} & \textbf{1.161} & \textbf{2.293} & \textbf{1.024} & \textbf{17.65} \\ \hline
        \multirow{2}{*}{Method} & \multicolumn{5}{c} {nuScenes Validation} \\ \cline{2-6}
        & 0-20 & 20-40 & 40-$\infty$ & Avg. & AP\\ \hline
        Baseline & 0.652 & 1.607 & 4.429 & 1.990 & 29.25 \\
        \textbf{Ours} & \textbf{0.547} & \textbf{1.500} & \textbf{3.661} & \textbf{1.669} & \textbf{32.21} \\  \hline 
        
    \end{tabular}}
    \vspace{0.2cm}
    \caption{Evaluation on KITTI validation and nuScenes frontal validation cars with depth MAE \textbf{($\downarrow$)} and AP \textbf{($\uparrow$)}.}
\label{tab:nus}
\end{table}

\subsubsection{Evaluation on nuScenes}
\label{sec:nus_exp}
It is noticed that recent semi-supervised works only evaluate KITTI~\cite{geiger2012kitti}. To further show the generalization and potential of our method, we conduct a toy experiment on nuScenes~\cite{nuscenes}. Knowing there is no established semi-supervised Mono3D prototype on nuScenes, we roughly divide the official training set of 28,130 images into two subsets: 3,375 labeled ones and 24,755 unlabeled ones. Evaluations are performed on the official validation set consisting of 6,019 images. Following the approach in \cite{kumar2022deviant}, we only consider the frontal view cars and use mean absolute error (MAE) of the depth and average precision (AP) to measure the prediction accuracy. We refer the readers to ~\cite{kumar2022deviant} for more details about the criteria. As shown in Tab.~\ref{tab:nus}, the proposed 3DSeMo shows consistent improvement across different distance ranges to ego car, again demonstrating our method's generation. We hope our attempt can drive more interest in semi-supervised 3D object detection.

\section{Conclusion and Limitation}

In this paper, we present the `Augment and Criticize' policies to construct a general framework for self-training-based semi-supervised monocular 3D object detection. The proposed APG aggregates predictions from different views of unlabeled images for robust label generation. The CRS adopts a learnable critical module to measure the reward of each pseudo sample and filter noisy ones to enhance model training. Extensive experiments and analyses demonstrate the effectiveness of our approach. 

We hope our work could enlighten more research on semi-supervised monocular 3D object detection. There exists one limitation that we leave to future work: As presented in Tab.~\ref{tab:label_unlabel_ratio}, detection performance grows with the unlabeled data volume yet it has not impressively plateaued. Naturally, restocking more unlabeled samples from other sources (\textit{e.g.} Waymo~\cite{waymo} and nuScenes~\cite{nuscenes}) could further enhance the detection methods. However, the domain gap in different sources may compromise the effectiveness of semi-supervised learning. In future work, we will devote more effort to mitigate this gap for exploiting more unlabeled data, which we believe can further facilitate semi-supervised Mono3D tasks.


%


{\small
\bibliographystyle{ieee_fullname}
\bibliography{egbib}
}

\end{document}